\documentclass[twoside,11pt]{article}

\usepackage[preprint]{jmlr2e}
\usepackage[utf8]{inputenc} 
\usepackage[T1]{fontenc}    
\usepackage{hyperref}       
\usepackage{url}   
\usepackage[ruled,longend,linesnumbered]{algorithm2e}
\usepackage{float}
\usepackage{multirow}
\usepackage{booktabs}       
\usepackage{amsfonts}    \usepackage{rotating}
\usepackage{nicefrac}       
\usepackage{microtype}      
\usepackage{microtype}
\usepackage{graphicx}
\usepackage{tabularx}
\usepackage{subfigure}
\usepackage[dvipsnames]{xcolor}
\usepackage{graphicx}
\usepackage{amsmath}
\usepackage{enumitem}
\usepackage{caption}
\usepackage{listings}
\usepackage[toc,header]{appendix}
\usepackage{minitoc}

\newcolumntype{C}{>{\centering\arraybackslash}X}

\hypersetup{
    colorlinks = true,
    allcolors = {Tan},
    linkbordercolor = {white},
}

\ShortHeadings{Temporally-Extended Prompts Optimization}{Chuyun Shen et al.}
\firstpageno{1}

\begin{document}

\doparttoc 
\faketableofcontents 

\title{Temporally-Extended Prompts Optimization for SAM in Interactive Medical Image Segmentation}

\author{\name Chuyun~Shen \email 
    cyshen@stu.ecnu.edu.cn \\
    \addr School of Computer Science and Technology\\
    East China Normal University\\
    Shanghai 200062, China
    \AND
    \name Wenhao~Li \email liwenhao@cuhk.edu.cn\\
    \addr School of Data Science\\
    The Chinese University of Hong Kong, Shenzhen\\
    Shenzhen Institute of Artificial Intelligence and Robotics for Society\\
    Shenzhen 518172, China
    \AND
    \name Ya~Zhang \email ya\_zhang@sjtu.edu.cn \\
    \addr Cooperative Medianet Innovation Center\\
    Shanghai Jiao Tong University\\
    Shanghai, 200240, China
    \AND
    \name Xiangfeng~Wang \email xfwang@cs.ecnu.edu.cn \\
    \addr School of Computer Science and Technology\\
    East China Normal University\\
    Shanghai 200062, China
}

\maketitle

\begin{abstract}
The \textit{Segmentation Anything Model} (SAM) has recently emerged as a foundation model for addressing image segmentation. 
Owing to the intrinsic complexity of medical images and the high annotation cost, the medical image segmentation (MIS) community has been encouraged to investigate SAM's zero-shot capabilities to facilitate automatic annotation. 
Inspired by the extraordinary accomplishments of \textit{interactive} medical image segmentation (IMIS) paradigm, this paper focuses on assessing the potential of SAM's zero-shot capabilities within the IMIS paradigm to amplify its benefits in the MIS domain. 
Regrettably, we observe that SAM's vulnerability to prompt forms (e.g., points, bounding boxes) becomes notably pronounced in IMIS. 
This leads us to develop a framework that adaptively offers suitable prompt forms for human experts. 
We refer to the framework above as \textit{temporally-extended prompts optimization} (TEPO) and model it as a Markov decision process, solvable through reinforcement learning. 
Numerical experiments on the standardized benchmark \texttt{BraTS2020} demonstrate that the learned TEPO agent can further enhance SAM's zero-shot capability in the MIS context.

\end{abstract}

\section{Introduction}\label{sec:intro}

The \textit{Segmentation Anything Model} (SAM)~\citep{sam} has recently been  proposed as a foundational model for addressing image segmentation problems.
SAM's effectiveness is principally evaluated in natural image domains, demonstrating a remarkable prompt-based, zero-shot generalization capability. 
Segmentation within medical images (MIS), on the other hand, presents complex challenges owing to their substantial deviation from natural images, encompassing multifaceted modalities, intricate anatomical structures, indeterminate and sophisticated object boundaries, and extensive object scales~\citep{sharma2010automated,hesamian2019deep,huang2023segment}.

Predominant MIS methods principally employ domain-specific architectures and necessitate reliance upon massive, high-quality expert annotations~\citep{unet,nnunet,nnformer,swin-unet}. 
In light of the considerable expenditure incurred by dense labeling, the community has embarked on exploring SAM's zero-shot generalization capabilities in MIS tasks, thereby fostering automated annotation of medical images~\citep{ji2023sam,ji2023segment,mohapatra2023brain,deng2023segment,zhou2023can,he2023accuracy,mazurowski2023segment,ma2023segment,cheng2023sam,zhang2023segment,roy2023sam,huang2023segment,mattjie2023exploring}.

Motivated by the remarkable achievements of \textit{interactive} medical image segmentation (IMIS), this paper goes a step further and centers on investigating the potential of zero-shot capabilities of SAM in the IMIS domain to magnify the advantages of SAM in MIS domain.
An extensive body of research demonstrates the significant performance enhancement attributable to the IMIS paradigm~\citep{xu2016deep,deepcut,scribblesup,polygon-rnn,deepigeos,seednet,itermrl,boundary-aware,mecca}.
Specifically, IMIS overcomes the performance limitation inherent in end-to-end MIS approaches by reconceptualizing MIS as a multi-stage, human-in-the-loop task.
At each iterative stage, medical professionals impart valuable feedback (e.g., designating critical points, demarcating boundaries, or construing bounding boxes) to identify inaccuracies in the model output. 
Consequently, the model refines the segmentation results following expert knowledge embedded in human feedback.

The congruity between the human feedback forms in IMIS and the prompt forms in SAM facilitates the seamless integration of SAM within the IMIS framework. 
Nevertheless, recent investigations reveal that, in contrast to natural image segmentation, the susceptibility of the SAM model to prompt forms (e.g., points or bounding boxes) is significantly heightened within MIS tasks, resulting in substantial discrepancies in zero-shot performance when various prompt forms are employed~\citep{cheng2023sam,roy2023sam,zhang2023segment}. 
Regrettably, we find this issue is markedly exacerbated within the IMIS context. 

This phenomenon can be attributed to two primary factors.
Firstly, the segmentation stages are interdependent; 
the previous prompt forms selection directly impacts the ensuing segmentation, which, in turn, influences the choice of subsequent prompt forms. 
Secondly, human experts display preferences and stochasticity in their feedback, seldom contemplating the ramifications of the prompt forms on the performance and the intricate interconnections between antecedent and successive prompt forms.
Consequently, this revelation impels us to recommend the most efficacious prompt forms for human feedback at each successive IMIS stage, a challenge we designate as \textit{temporally-extended prompts optimization}.

As a formidable instrument for addressing sequential decision-making, reinforcement learning (RL)~\citep{sutton-rl} demonstrates remarkable competencies not only in domains such as chess, video games, and robotics control but also in training foundational models~\citep{instructGPT,wei2022emergent} and IMIS~\citep{itermrl,boundary-aware,mecca}.
Given that temporally-extended prompt optimization encompasses both the foundational model and IMIS, we formulate this problem as a Markov decision process (MDP) and employ RL for its resolution.
The framework above is then instantiated as the algorithm denoted by \textbf{TEPO}.
During each stage, TEPO agent determines which prompt form is most suitable for recommendation to human, considering the current segmentation outcomes and historical prompts.
The ultimate objective is to augment the performance of the SAM in each stage relative to its preceding iteration, thereby maximizing its efficacy.

The contributions presented in this paper encompass three distinct aspects:
1) In an unprecedented discovery, we ascertain that sequential prompt forms constitute the crucial elements influencing the zero-shot performance of SAM in IMIS, subsequently proposing a pertinent temporally-extended prompts optimization problem;
2) By conceptualizing the temporally-extended prompts optimization as an MDP, we employ RL to optimize the sequential selection of prompt forms, thereby enhancing the zero-shot performance of SAM in IMIS;
3) The performance juxtaposition and ablation studies conducted on the standardized benchmark \texttt{BraTS2020} substantiate the efficacy of the TEPO agent in ameliorating SAM's zero-shot capability.

\section{Related Work and Preliminaries}

\subsection{Interactive Medical Image Segmentation}

Before the remarkable advancements in automatic segmentation achieved through convolutional neural networks (CNNs), traditional interactive techniques were employed within IMIS~\citep{zhao2013overview}. Among these techniques, the RandomWalk method~\citep{grady2006random} generates a weight map with pixels as vertices and segments images based on user interaction. 
Approaches like GrabCut~\citep{rother2004grabcut} and GraphCut~\citep{boykov2001interactive} establish a connection between image segmentation and graph theory's maximum flow and minimum cut algorithms. Geos~\citep{criminisi2008geos} introduces a geodesic distance measurement to ascertain pixel similarity.

There has been a surge of interest in deep learning-based IMIS methods in recent years. 
\citet{xu2016deep} suggests employing CNNs for interactive image segmentation, whereas DeepCut~\citep{deepcut} and ScribbleSup~\citep{scribblesup} utilize weak supervision in developing interactive segmentation techniques. 
DeepIGeoS~\citep{deepigeos} incorporates a geodesic distance metric to generate a hint map.

The interactive segmentation process can be viewed as a sequential procedure, which makes it a natural fit for reinforcement learning (RL).
Polygon-RNN~\citep{polygon-rnn} tackles this problem by segmenting targets as polygons and iteratively selecting polygon vertices through a recurrent neural network (RNN). 
With a similar approach, Polygon-RNN+~\citep{polygon-rnn++} adopts a similar approach to Polygon-RNN, it employs RL to learn vertex selection. 
SeedNet~\citep{seednet} takes a different approach by constructing an expert interaction generation RL model that can obtain simulated interaction data at each interaction stage.

IteR-MRL~\citep{itermrl} and BS-IRIS~\citep{boundary-aware} conceptualize the dynamic interaction process as a Markov Decision Process (MDP) and apply multi-agent RL models for image segmentation purposes. 
MECCA~\citep{mecca}, based on IteR-MRL, establishes a confidence network, seeking to mitigate the pervasive ``interactive misunderstanding'' issue that plagues RL-based IMIS techniques and ensure the effective utilization of human feedback.
Additionally, \citet{liu2023samm} integrates the SAM within the \textit{3D Slicer} software, thereby facilitating the process of designing, evaluating, and employing SAM in the context of IMIS.

\subsection{Segment Anything Model}

The \textit{Segmentation Anything Model} (SAM)~\citep{sam}, recently introduced by Meta, serves as a fundamental framework for tackling image segmentation challenges. 
Motivated by the robust performance of foundational models in NLP and CV domains, researchers endeavored to establish a unified model for complete image segmentation tasks. 
Nonetheless, the actual data in the segmentation field necessitates revision and diverges from the design intentions mentioned above.
Consequently, \citet{sam} stratifies the process into three distinct phases: \textit{task}, \textit{model}, and \textit{data}. 
Refer to the primary publication~\citep{sam} and a contemporary survey~\citep{zhang2023comprehensive} for comprehensive explanations.

\smallskip
\noindent \textbf{Task.}
Drawing inspiration from foundational NLP and CV models, \citet{sam} introduces the promptable segmentation task to generate a valid segmentation mask in response to any given segmentation prompt. 
These prompts define the target object(s) to be segmented within an image and may include a location point, a bounding box, or a textual description of the object(s). 
The resulting mask must be plausible for at least one target object, even in instances where the prompt may be ambiguous or reference multiple objects.

\begin{figure}[htb!]
    \centering
    \includegraphics[width=\linewidth]{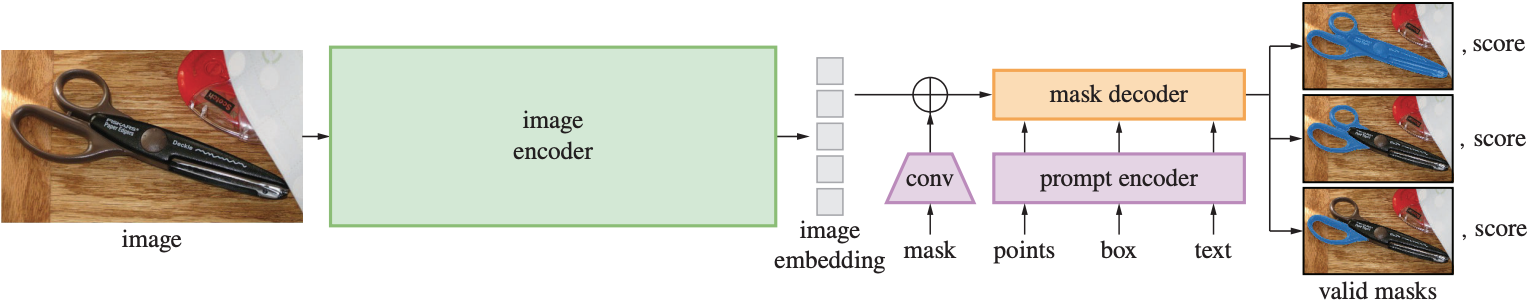}
    \caption{The screenshot of SAM~\citep{sam}.}
    \label{fig:sam}
\end{figure}

\smallskip
\noindent \textbf{Model.} 
The promptable segmentation task, paired with the objective of real-world applicability, imposes restrictions on the model architecture. 
\citet{sam} devises a streamlined yet efficacious model, known as SAM (Figure~\ref{fig:sam}), which encompasses a powerful image encoder that computes image embeddings, a prompt encoder that embeds prompts, and a lightweight mask decoder that amalgamates the two information sources to predict segmentation masks.

\smallskip
\noindent \textbf{Data.}
SAM necessitates training on an extensive and diverse collection of masks to attain exceptional generalization capabilities on novel data distributions. 
\citet{sam} constructs a "data engine", employing a model-in-the-loop dataset annotation approach, thereby co-developing SAM in tandem. 
The resulting dataset, SA-1B, incorporates over $1$ billion masks derived from $11$ million licensed and privacy-preserving images.

\subsection{Segment Anything in Medical Images}

Building upon the foundational pre-trained models of SAM, many papers have delved into investigating its efficacy in diverse zero-shot MIS scenarios. 
\citet{ji2023sam} conducts a comprehensive evaluation of SAM in the \textit{everything} mode for segmenting lesion regions within an array of anatomical structures (e.g., brain, lung, and liver) and imaging modalities (\textit{computerized tomography}, abbreviated as CT, and \textit{magnetic resonance imaging}, abbreviated as MRI).
\citet{ji2023segment} subsequently scrutinizes SAM's performance in specific healthcare domains (optical disc and cup, polyp, and skin lesion segmentation) utilizing both the automatic \textit{everything} mode and the manual \textit{prompt} mode, employing points and bounding boxes as prompts.

For MRI brain extraction tasks, \citet{mohapatra2023brain} compares SAM's performance with the renowned \textit{Brain Extraction Tool} (BET), a component of the \textit{FMRIB Software Library}.
\citet{deng2023segment} appraises SAM's performance in digital pathology segmentation tasks, encompassing tumor, non-tumor tissue, and cell nuclei segmentation on high-resolution whole-slide imaging.
\citet{zhou2023can} adeptly implements SAM in polyp segmentation tasks, utilizing $5$ benchmark datasets under the \textit{everything} setting.
Recently, an assortment of studies has rigorously tested SAM on over $10$ publicly available MIS datasets or tasks~\citep{he2023accuracy, mazurowski2023segment, ma2023segment, wu2023medical, huang2023segment, zhang2023customized}.

Quantitative experimental results gleaned from these works reveal that the zero-shot performance of SAM is, on the whole, moderate and exhibits variability across distinct datasets and cases.
To elaborate:
1) Utilizing \textit{prompt} instead of \textit{everything} mode, SAM can surpass state-of-the-art (SOTA) performance in tasks characterized by voluminous objects, smaller quantities, and well-defined boundaries when reliant on dense human feedback;
2) However, a considerable performance discrepancy remains between SAM and SOTA methods in tasks involving dense and amorphous object segmentation.

\section{Temporally-extended Prompts Optimization Methodology}

\begin{figure}[htb!]
    \centering
    \includegraphics[width=0.6\linewidth]{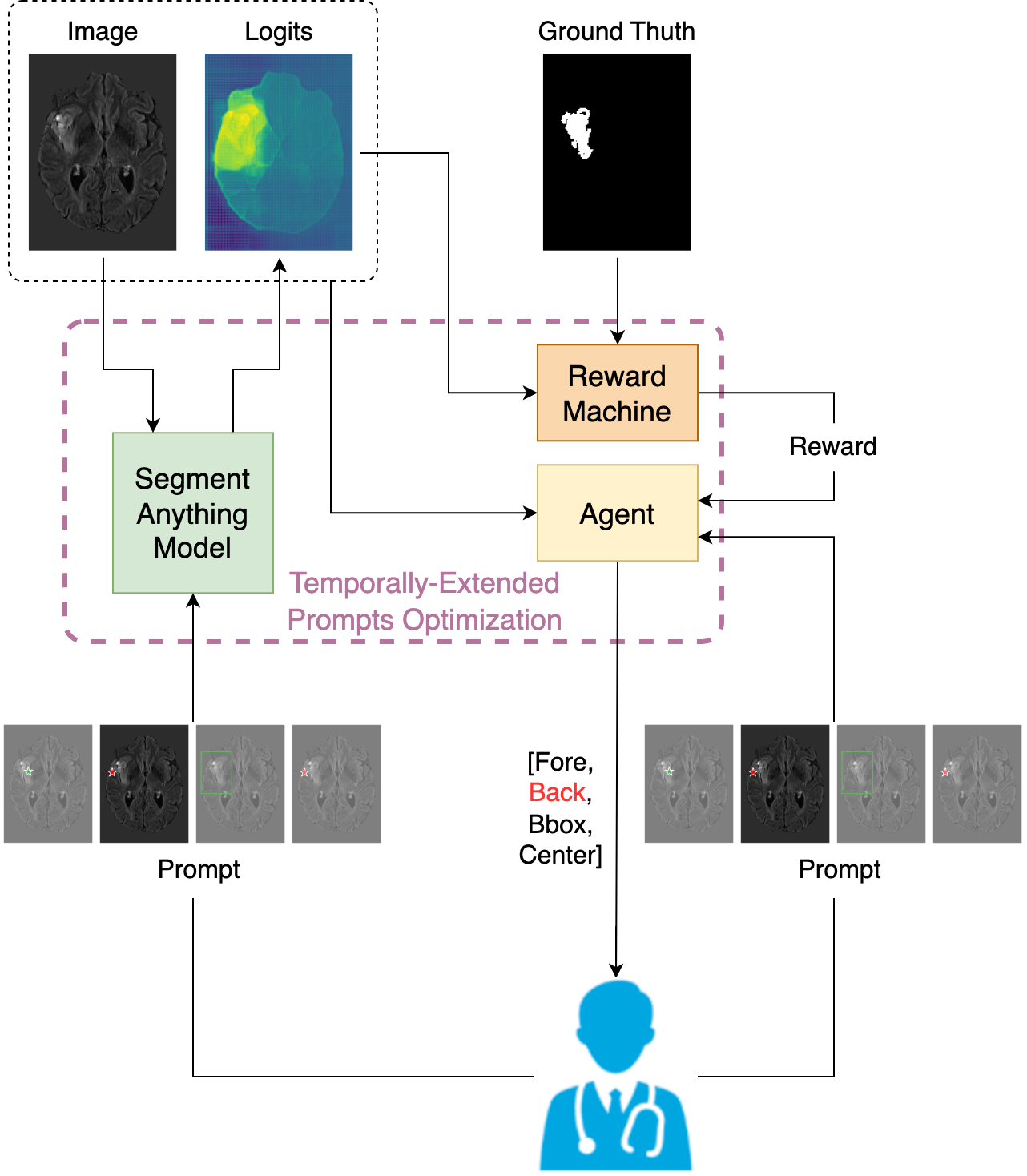}
    \caption{The architecture of our proposed TEPO.}
    \label{fig:TEPO}
\end{figure}

As elucidated in the preceding analysis, the susceptibility of SAM to prompt forms is markedly pronounced in IMIS. 
This serves as the impetus for devising a framework adept at adaptively proffering suitable prompt forms for human specialists, contingent upon the current progression of segmentation. 
The human expert subsequently imparts feedback to SAM, employing the recommended prompt form. 
The ensuing discourse delineates the modeling of this framework, the temporally-extended prompts optimization, as an MDP (Section~\ref{sec:mdp}) and elaborates on its solution through reinforcement learning (Section~\ref{sec:tepo}).

\subsection{Problem Formulation}\label{sec:mdp}

We consider a standard setup consisting of an agent interacting with an environment in discrete timesteps.
In our setting, the purpose of the agent is to recommend appropriate prompt forms for human experts.
At each timestep $t$ the agent receives an observation $o_t$, takes an action $a_t$ and receives a scalar reward $r_t$. 
In general, the environment may be partially observed so that the entire history of the observations, action pairs $s_t=\left(o_1, a_1, \ldots, a_{t-1}, o_t\right)$ may be required to describe the state.

The behavior of an agent is defined by a policy, $\pi$, which maps states to a probability distribution over actions $\pi: \mathcal{S} \rightarrow \mathcal{P}(\mathcal{A})$. 
We model it as a Markov decision process with state space $\mathcal{S}$, action space $\mathcal{A}$, initial state distribution $p\left(s_1\right)$, transition dynamics $p\left(s_{t+1} \mid s_t, a_t\right)$, reward function $r\left(s_t, a_t\right)$, and instantiate it as follows:

\smallskip
\noindent\textbf{State space.}
The state at timestep $t$ is represented as a three-tuple $S_t = (I, P_{t-1}, T_{t-1})$, where $I \in \mathbb{R}^{H\times W\times C}$ represents the medical image slice input, $P_{t-1} \in \mathbb{R}^{H\times W\times K}$ represents the segmentation logits from the previous time step $t-1$ (where $K$ represents the number of segmentation classes, which in this case is 2), and $T_{t-1}$ is a set of interaction prompts provided before time step $t$.
We consider four types of interaction prompts form at each timestep: forehead, background, center point, and bounding box, and we will introduce them in Section~\ref{sec:results}.

\smallskip
\noindent\textbf{Action space.}
The action space $\mathcal{A}$ is a set of interactive forms provided by human experts at each time step. It is represented as a set of integers $\mathcal{A}=\{0,1,2,3\}$, where $0$ denotes selecting the forehead point, $1$ denotes accessing the background point, $2$ denotes the center point, which is defined as the point farthest from the boundary of the error regions, and $3$ denotes selecting the bounding box.
At each time step, the agent chooses an action from the action space $\mathcal{A}$ to assist human experts with their interactions with SAM.

\smallskip
\noindent\textbf{Reward function.}
At each step t, the difference between the current DICE score~\citep{dice1945measures}, $dice(P_t, Y)$ and the previous DICE score, $dice(P_{t-1}, Y)$, is calculated as the reward value $R_t$:
$$R_t = dice(P_t, Y) - dice(P_{t-1}, Y),$$where $Y$ is the ground truth, $dice(P_t, Y)$ represents the DICE score between the current predicted result $P_t$ and the ground truth, and $dice(P_{t-1}, Y)$ represents the DICE score between the previous predicted result and the ground truth. 

In summary, as shown in Figure \ref{fig:TEPO}, the whole process is as follows:
the agent gives the recommended prompt form in accordance with the policy $pi$ based on the raw image, the current segmentation probability and the hints given by the doctor.
Then the doctor gives the SAM the corresponding prompt and updates the segmentation probability.
The changes in the segmentation result are used by the agent as the reward to update the policy $\pi$.

In addition, the \textit{return} from a state is defined as the sum of discounted future reward $R_t=\sum_{i=t}^T \gamma^{(i-t)} r\left(s_i, a_i\right)$ with a discounting factor $\gamma \in[0,1]$.
Depending on this problem formulation, the goal of temporally-extended prompts optimization is to learn a policy that maximizes the expected return from the start distribution $J=\mathbb{E}_{r_i, s_i \sim E, a_i \sim \rho^\pi}\left[R_1\right]$. 
We denote the discounted state visitation distribution for a policy $\pi$ as $\rho^\pi$.

\subsection{Learning the TEPO Agent with RL}\label{sec:tepo}

Before introducing RL methods to obtain the optimal prompt, we first introduce some notations.
The \textit{action-value function} is used in many RL algorithms. 
It describes the expected return after taking an action $a_t$ in state $s_t$ and thereafter following policy $\pi$ :
\begin{equation}
    Q^\pi\left(s_t, a_t\right)=\mathbb{E}_{r_{i \geq t}, s_{i>t} \sim E, a_{i>t} \sim \pi} \big[ R_t \mid s_t, a_t \big].
\end{equation}

Additionally, many approaches in RL make use of the recursive relationship known as the \textit{Bellman equation}:
\begin{equation}
        Q^\pi \left( s_t, a_t \right) = \mathbb{E}_{r_t, s_{t+1} \sim E} \left[ r \left( s_t, a_t \right) + \gamma \mathbb{E}_{a_{t+1} \sim \pi} \big[ Q^\pi \left( s_{t+1}, a_{t+1} \right) \big] \right].
\end{equation}
This paper adopts deep $Q$-network (DQN)~\citep{dqn1,dqn2} to instantiate the RL framework and learn the TEPO agent.
$Q$-learning~\citep{qlearning}, as the core module of DQN, is a commonly-used, off-policy RL algorithm, by employing the greedy policy $\mu(s)=\arg \max _a Q(s, a)$.
DQN adapts the $Q$-learning to make effective use of large neural networks as action-value function approximators.
% To scale $Q$-learning, DQN introduces two major changes: the use of a replay buffer, and a separate target network for calculating $y_t$.
We consider function approximators parameterized by $\theta^Q$, which we optimize by minimizing the loss:
\begin{equation}
\!\!\!\! {\cal{L}}\left(\theta^Q\right)=\mathbb{E}_{s_t \sim \rho^\beta, a_t \sim \beta, r_t \sim E}\left[\left(Q\left(s_t, a_t \mid \theta^Q\right)-y_t\right)^2\right],
\end{equation}
where $y_t=r\left(s_t, a_t\right)+\gamma Q\left(s_{t+1}, \mu\left(s_{t+1}\right) \mid \theta^Q\right)$.
The full algorithm, which we call TEPO, is presented in Algorithm \ref{algorithm}.

\begin{algorithm}[htb!]
    \caption{Temporally-Extended Prompts Optimization}\label{algorithm}
    Initialize replay memory $\mathcal{D}$ to capacity $N$ \\
    Initialize action-value function $Q$ with random weights \\
    \For{episode $=1, M$}{
        Randomly sample an image $I$, and encode it with SAM \\
        Initialise state $s_{1}=\left( I, P_0
    , T_0 \right)$, where $P_0 = \textbf{0}$ and $T_0 = \emptyset$ \\
        \For{$t=1,T$}{
            With probability $\epsilon$ select a random avail interaction form $a_{t}$; otherwise select $a_{t}=\max _{a} Q^{*}\left(s_{t}, a ; \theta\right)$ \\
            Execute action $a_{t}$ and doctor make a prompt based on $a_{t}$ and $s_{t}$, and then SAM input prompts $T_t$ and update logit $P_t$ \\
            Reward $r_{t} = r\left(s_t, a_t\right)$  and update state $s_{t+1} = \left( I, P_t, T_t \right) $ \\
            Store transition $ \left( s_{t}, a_{t}, r_{t}, s_{t+1} \right)$ in $\mathcal{D}$ \\
            Sample random minibatch of transitions $\left(s_{j}, a_{j}, r_{j}, s_{j+1}\right)$ from $\mathcal{D}$ \\
            Set greedy action $\mu(s)=\arg \max _a Q(s, a)$ \\
            Set $y_{j}=\left\{\begin{array}{ll}r_{j} &  \textrm{for terminal $s_{j+1}$}   \\ 
            r_{j}+\gamma Q\left(s_{j+1}, \mu(s_{j+1}) \mid \theta^Q \right) &  \textrm{for non-terminal $s_{j+1}$}   \end{array}\right.$ \\
            Perform gradient descent based on the Bellman error $\left(Q\left(s_t, a_t \mid \theta^Q\right)-y_t\right)^2$ 	
            }			
    }
\end{algorithm}

\section{Experiments on MIS}

This section provides an evaluation of the proposed TEPO on the \texttt{BraTS2020} benchmark, which is a prevalent dataset used for MIS tasks. We aim to address the following key questions, and the following evaluation will focus on answering these questions comprehensively, i.e.,

\smallskip
\noindent {{a) Does the SAM with multi-step interaction outperform the SAM with single-step interaction?}

\noindent b) Can the policies learned by the TEPO algorithm outperform the rule-based policies?

\noindent c) What strategies can be learned from TEPO?

\noindent d) How stable are the strategies learned by TEPO?

\subsection{Dataset and Training Details}

SAM requires 2D images as input and 3D images are conventionally often annotated by viewing them in slices, we adopt the practice of slicing the 3D magnetic resonance scans into axial slices, a method commonly used in related research efforts~\citep{wolleb2022diffusion}.

To evaluate the effectiveness of TEPO in the context of multi-step interaction, we carefully selected slices with sufficiently large foregrounds in the image. 
Specifically, we segment the \textit{Whole Tumor} (WT) from the \textit{FLAIR} images and choose slices that contain a minimum of 256 foreground pixel points for analysis. This carefully curated dataset enables accurate evaluation of the performance and potential of TEPO in future applications in MIS.

The dataset for evaluation comprised a total of 369 patients. 
We split the dataset into three subsets: the training set evaluated $319$ patients and included $17,396$ slices; the validation set consisted of $20$ patients, corresponding to $1,450$ slices; and the test set included $20$ patients with $1,389$ slices.

We crop the images to $200 \times 150$, implement \textit{random flip}, \textit{rotate}, \textit{add noise}, \textit{affine transform} data augmentation to the training dataset, and then rescale the intensity values.
We train for $100$ epochs, and in each epoch, $10,000$ steps are sampled, and the $Q$ network of TEPO is updated $100$ times.
The model is trained with a learning rate of $1e^{-3}$ for the \texttt{Adam} optimizer and a batch size of $64$.

\subsection{Main Results}\label{sec:results}

\begin{sidewaystable} [!htp]
\caption{Action selection preference statistics and quantitative segmentation performance results for TEPO policies and rule-based policies. Labels used in the paper include ``Fore'' for the forehead point form, ``Back'' for the background point form, ``Center'' for the center point form, and ``Bbox'' for the bounding box form. These labels will be consistently used throughout the paper. 
``<-0.1" indicates the number of cases that reward less than 0.1, which means the algorithm misunderstands the interaction.  
In addition, we use boldface to indicate the highest dice score in each step.}
\label{tab:tepo_tab}
\centering
\resizebox{\linewidth}{!}{
\begin{tabular}{|l|l|l|l|l|l|l|l|l|l|l|} 
\hline
Algorithm & variable & Step 1 & Step 2 & Step 3 & Step 4 & Step 5 & Step 6 & Step 7 & Step 8 & Step 9 \\ 
\hline
\multirow{3}{*}{TEPO-2} & Action & Bbox (100.00\%) & Fore (99.57\%) & Fore (99.78\%) & Fore (99.71\%) & Fore (99.93\%) & Fore (99.86\%) & Fore (99.86\%) & Fore (99.86\%) & Fore (99.93\%) \\ 
\cline{2-11}
 & Dice & 0.\textbf{6901$ \pm $ 0.2094} & 0.6930$ \pm $ 0.1758 & 0.6937$ \pm $ 0.1694 & 0.6932$ \pm $ 0.1692 & 0.6940$ \pm $ 0.1693 & 0.6940$ \pm $ 0.1693 & 0.6940$ \pm $ 0.1694 & 0.6940$ \pm $ 0.1694 & 0.6940$ \pm $ 0.1694 \\ 
\cline{2-11}
 &  -0.1 & 0 & 95 & 14 & 2 & 0 & 0 & 0 & 0 & 0 \\ 
\hline
\multirow{3}{*}{TEPO-3} & Action & Center (100.00\%) & Bbox (94.96\%) & Center (98.85\%) & Center (99.14\%) & Center (99.78\%) & Center (100.00\%) & Center (99.86\%) & Center (100.00\%) & Center (100.00\%) \\ 
\cline{2-11}
 & Dice & 0.4658$ \pm $ 0.2877 & \textbf{0.7035$ \pm $ 0.1882} & \textbf{0.7611$ \pm $ 0.1687} & \textbf{0.7845$ \pm $ 0.1670} & \textbf{0.8026$ \pm $ 0.1553} & \textbf{0.8198$ \pm $ 0.1441} & 0.8263$ \pm $ 0.1409 & 0.8332$ \pm $ 0.1367 & 0.8362$ \pm $ 0.1378 \\ 
\cline{2-11}
 &  -0.1 & 0 & 54 & 44 & 72 & 73 & 46 & 47 & 45 & 44 \\ 
\hline
\multirow{3}{*}{TEPO-5} & Action & Center (100.00\%) & Center (62.35\%) & Center (86.54\%) & Center (95.10\%) & Center (97.48\%) & Center (99.57\%) & Center (99.64\%) & Center (99.71\%) & Center (99.64\%) \\ 
\cline{2-11}
 & Dice & 0.4658$ \pm $ 0.2877 & 0.6472$ \pm $ 0.2316 & 0.7369$ \pm $ 0.1926 & 0.7782$ \pm $ 0.1665 & 0.8021$ \pm $ 0.1577 & 0.8190$ \pm $ 0.1439 & 0.8288$ \pm $ 0.1375 & 0.8372$ \pm $ 0.1346 & 0.8421$ \pm $ 0.1322 \\ 
\cline{2-11}
 &  -0.1 & 0 & 117 & 102 & 103 & 85 & 62 & 48 & 44 & 42 \\ 
\hline
\multirow{3}{*}{TEPO-7} & Action & Center (100.00\%) & Bbox (85.67\%) & Center (79.34\%) & Center (90.78\%) & Center (93.38\%) & Center (94.89\%) & Center (95.39\%) & Center (95.82\%) & Center (95.61\%) \\ 
\cline{2-11}
 & Dice & 0.4658$ \pm $ 0.2877 & 0.6981$ \pm $ 0.1965 & 0.7552$ \pm $ 0.1720 & 0.7822$ \pm $ 0.1690 & 0.7991$ \pm $ 0.1612 & 0.8137$ \pm $ 0.1520 & 0.8240$ \pm $ 0.1424 & 0.8316$ \pm $ 0.1370 & 0.8342$ \pm $ 0.1380 \\ 
\cline{2-11}
 &  -0.1 & 0 & 56 & 56 & 65 & 62 & 53 & 40 & 45 & 41 \\ 
\hline
\multirow{3}{*}{TEPO-9} & Action & Center (100.00\%) & Center (100.00\%) & Center (100.00\%) & Center (100.00\%) & Center (100.00\%) & Center (100.00\%) & Center (100.00\%) & Center (100.00\%) & Center (100.00\%) \\ 
\cline{2-11}
 & Dice & 0.4658$ \pm $ 0.2877 & 0.6211$ \pm $ 0.2535 & 0.7192$ \pm $ 0.2131 & 0.7707$ \pm $ 0.1711 & 0.7990$ \pm $ 0.1583 & 0.8175$ \pm $ 0.1452 & \textbf{0.8302$ \pm $ 0.1390} & \textbf{0.8394$ \pm $ 0.1324} & \textbf{0.8449$ \pm $ 0.1307} \\ 
\cline{2-11}
 &  -0.1 & 0 & 177 & 153 & 167 & 123 & 86 & 58 & 51 & 51 \\ 
\hline
\multirow{3}{*}{Random} & Action & Center (26.78\%) & Center (29.30\%) & Fore (30.67\%) & Fore (33.48\%) & Back (32.04\%) & Fore (33.48\%) & Center (33.33\%) & Back (33.33\%) & Fore (33.55\%) \\ 
\cline{2-11}
 & Dice & 0.4129 $ \pm $ 0.3417 & 0.5723 $ \pm $ 0.2947 & 0.6561 $ \pm $ 0.2562 & 0.7072 $ \pm $ 0.2260 & 0.7354 $ \pm $ 0.2094 & 0.7571 $ \pm $ 0.1943 & 0.7818 $ \pm $ 0.1727 & 0.7956 $ \pm $ 0.1627 & 0.8052 $ \pm $ 0.1568 \\ 
\cline{2-11}
 &  -0.1 & 0 & 121 & 141 & 123 & 115 & 89 & 66 & 50 & 39 \\ 
\hline
\multirow{3}{*}{Alternately} & Action & Fore (100.00\%) & Back (100.00\%) & Fore (100.00\%) & Back (100.00\%) & Fore (100.00\%) & Back (100.00\%) & Fore (100.00\%) & Back (100.00\%) & Fore (100.00\%) \\ 
\cline{2-11}
 & Dice & 0.4658 $ \pm $ 0.2877 & 0.6010 $ \pm $ 0.2691 & 0.6460 $ \pm $ 0.2470 & 0.7280 $ \pm $ 0.2067 & 0.7332 $ \pm $ 0.2098 & 0.7823 $ \pm $ 0.1730 & 0.7840 $ \pm $ 0.1777 & 0.8138 $ \pm $ 0.1512 & 0.8052 $ \pm $ 0.1652 \\ 
\cline{2-11}
 &  -0.1 & 0 & 98 & 207 & 98 & 172 & 59 & 111 & 27 & 92 \\
\hline
\end{tabular}
}
% \end{table}
\end{sidewaystable}

The performance of the proposed TEPO algorithm is evaluated on the \texttt{BraTS2020} dataset for medical image segmentation tasks and compared with three rule-based policy baselines: the one-step oracle agent, the random agent, and the alternately changing agent.
The one-step oracle agent is an optimal decision-making agent that has access to comprehensive information and can observe the reward after adapting various interaction forms.
This allows it to achieve the highest accuracy in a single step and to explore efficient interaction strategies for the given task.
The random agent, on the other hand, uniformly samples actions from available action sets and can be used to simulate clinicians without any preference for any particular interaction form for the task at hand.
The alternately changing agent applies a policy that alternately chooses the forehead point and the background point.

We evaluate the agent's performance through the dice score, computed using a ground truth mask and measurements taken at multiple timesteps ($N=\{2,3,5,7,9\}$).
At each timestep, the agent first chooses an action to indicate what form of interaction is required. 
To simulate a clinician's behavior, we use rules consisting of choosing specific positions, such as the forehead, background, and center, and drawing bounding boxes around the forehead region.
Specifically, we select the forehead, background, and center points that are farthest from the boundaries of the false negative, false positive, and error regions, respectively. 
For the bounding box, we extend the forehead region by $10$ pixels and draw a rectangle.

\begin{figure}[htb!]
  \centering
  \subfigure[Complete view.]{
    \label{fig:subfig1}
    \includegraphics[width=0.45\textwidth]{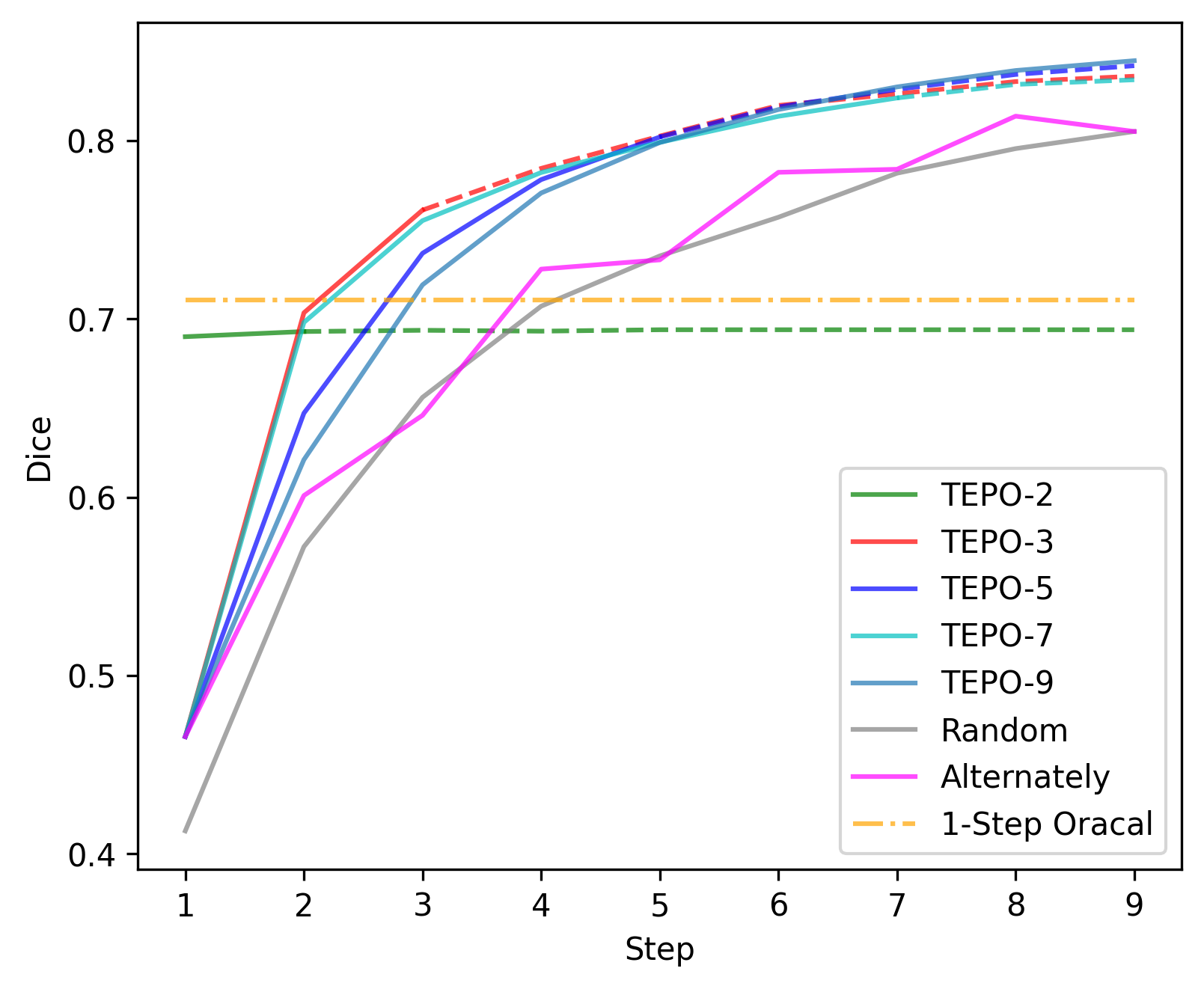}}
  \subfigure[Enlarged view.]{
    \label{fig:subfig2}
    \includegraphics[width=0.45\textwidth]{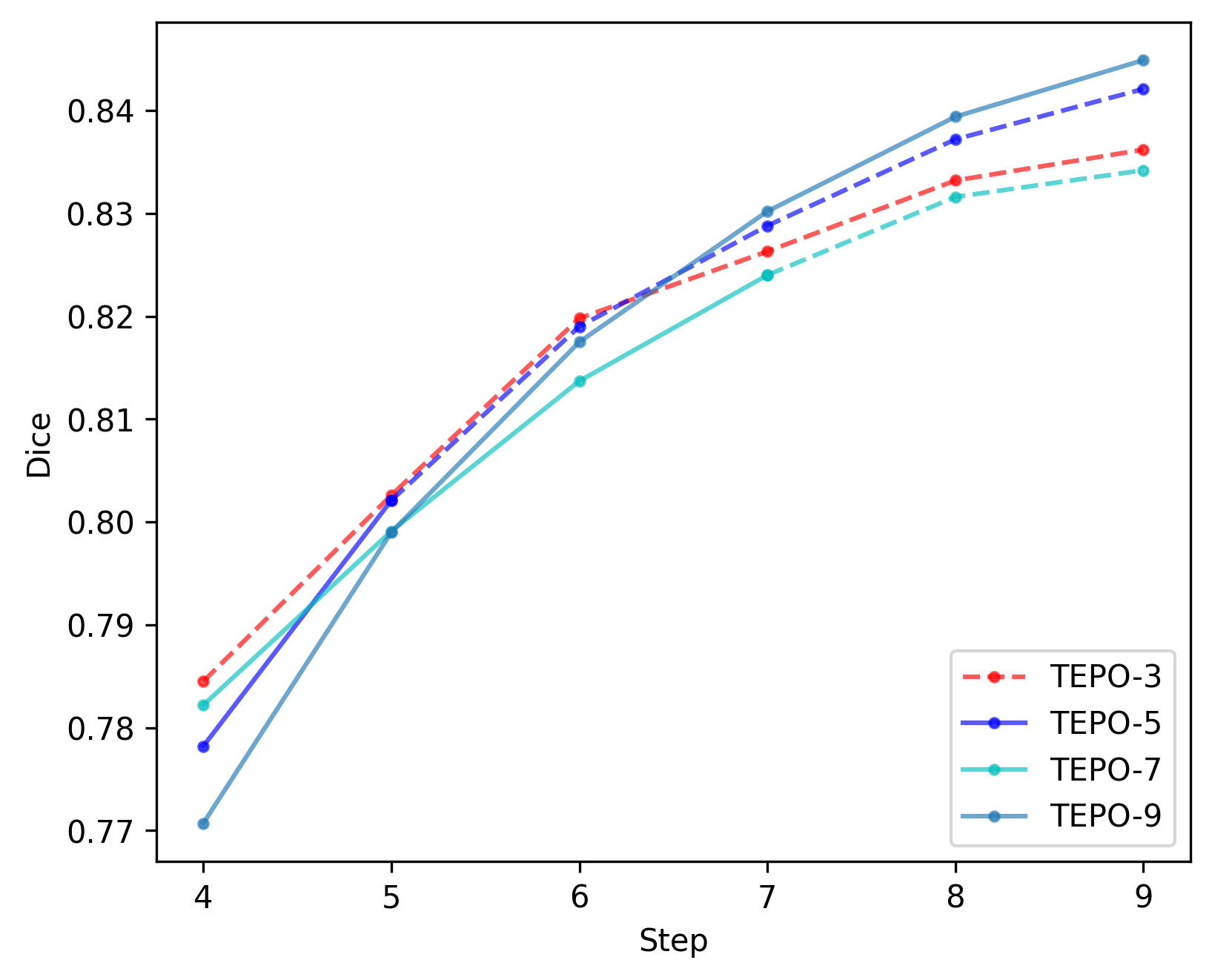}}
  \caption{The performance improvement of different IMIS methods at different interaction steps. All these test results are performed on the \texttt{BraTS2020} dataset. Subfigure (a) illustrate various IMIS methods from step one to step nine. To compare the performances of the algorithms, Subfigure (b) shows an enlarged view of steps four to nine, only including those with dice scores larger than 0.75.}
    \label{fig:dice}
\end{figure}

\begin{figure*}[htb!]
    \centering
    \includegraphics[width=0.9\linewidth]{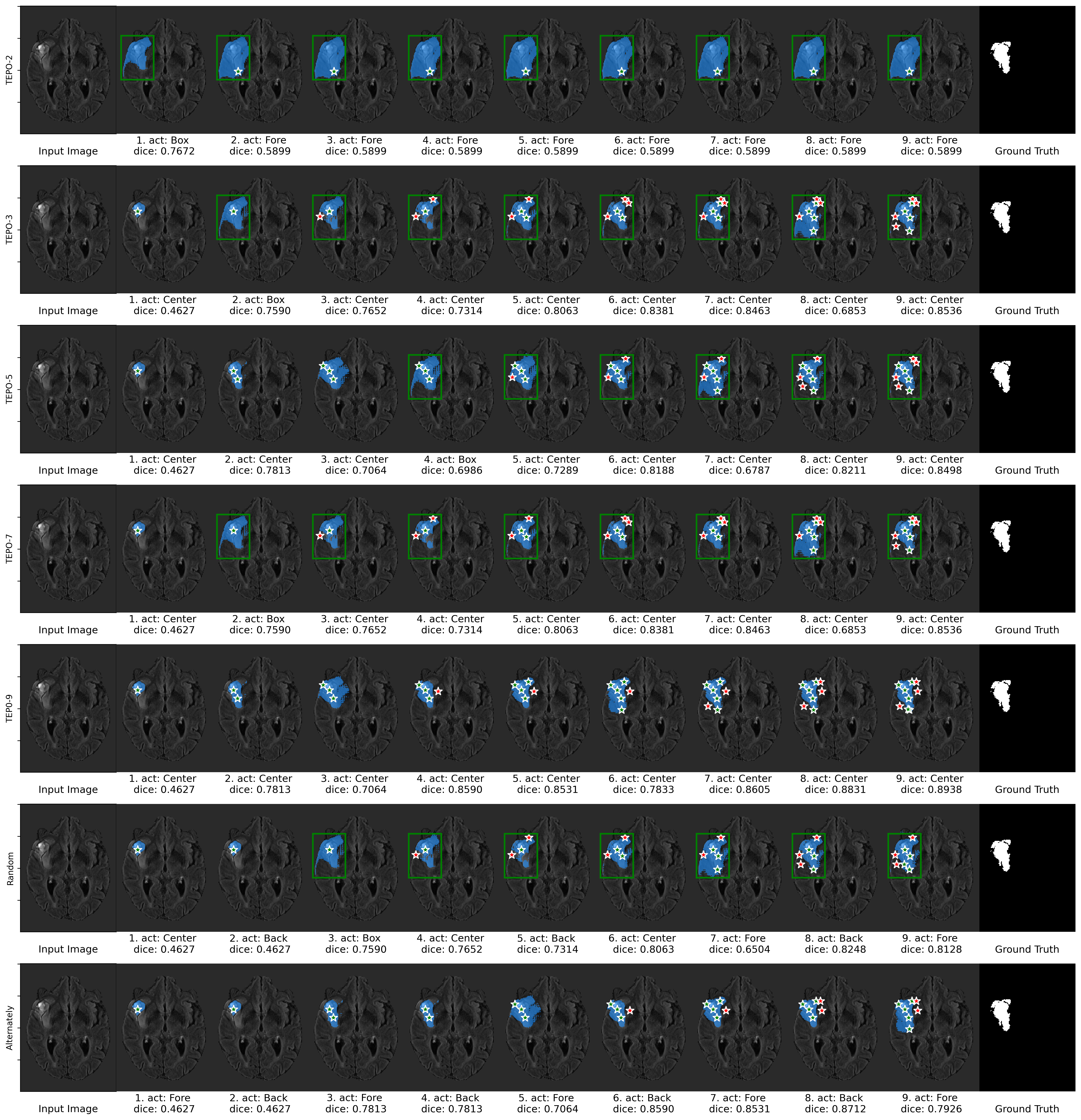}
    \caption{Visualization of strategies and results of different strategies on the same medical image. The green pentagram indicates the foreground, the red pentagram indicates the background, and the green box indicates the bounding box.}
    \label{fig:case_study}
\end{figure*}

The comparison of the performance of various interaction strategies is evaluated with respect to the number of interactions.
As shown in Figure \ref{fig:dice}, the different lines correspond to the different agents' performance.
``TEPO-X'' indicates that the agent is trained in the $X$-step interaction scenario. 
For example, ``TEPO-$2$'' means a two-step scenario. 
``Random'' denotes the random agent, ``Alternately'' denotes the alternately changing agent, and ``$1$-step Oracal'' denotes the one-step Oracal agent.
We will use the same labeling convention throughout the paper unless noted otherwise.
It is worth noting that we train in different interaction step scenarios, but in testing, we use $9$-step interactions to find out comprehensive performances.

\subsubsection{Quantitative experimental analysis}

\paragraph{Q\#a: Does the SAM in multi-step interaction mode outperform the SAM in single-step interaction mode?} 
As illustrated in Figure \ref{fig:dice}, the TEPO-$2$ agent stays the same after the third round, this is because in our experiments, if the shortest distance of all points from the edge in the corresponding region is less than two pixels, then the user does not interact anymore.
Table \ref{tab:tepo_tab} indicates that the TEPO-$2$ policy predominantly selects the forehead point starting from step two. 
However, the false negative region is too small to click, so the TEPO-$2$ policy stops interacting at step five for all test cases.
Conversely, the performance of other multi-step policies improves with an increase in the number of interactions, showcasing that SAM can be enhanced through multiple rounds of interactions.
Moreover, expect TEPO-$2$, other policies perform better than the one-step Oracle agent, implying that multi-step interactions are more effective for medical image segmentation than the single-step interaction mode.

\paragraph{Q\#b: Can the policies learned by the TEPO algorithm outperform the rule-based policies?}
The experimental results in Figure \ref{fig:dice} indicate that the TEPO-$2$ policy performs better than random and alternating selection methods during the initial two interactions. Moreover, the performance of all other RL-based policies is superior to rule-based approaches. 
These findings provide evidence that the TEPO algorithm significantly boosts the efficacy of SAM in interactive medical scenarios, even in zero-shot mode.

\paragraph{Q\#c: What strategies can be learned from the TEPO algorithm?}
As the TEPO algorithm is trained under different interaction round scenarios, the learned strategies exhibit variations, as summarized in Table \ref{tab:tepo_tab}.
TEPO-$2$ employs a straightforward strategy: selecting the bounding box in the first step and the forehead point in subsequent ones. 
This strategy performs well in the initial two steps, with the performance in the first step nearing that of the one-step Oracle agent that adopts an ideal strategy.
TEPO-$3$ applies a nearly deterministic strategy that chooses the bounding box at the second step and chooses center points at other steps.
Moreover, TEPO-$5$ and TEPO-$7$ use more uncertain strategies that primarily employ the center point but may resort to alternative ones in the second and third steps.
TEPO-$9$ finds a trivial strategy of choosing the center point at each step, resulting in the best performance in multiple interactions.

\begin{figure}[htb!]
    \centering
    \includegraphics[width=0.9\linewidth]{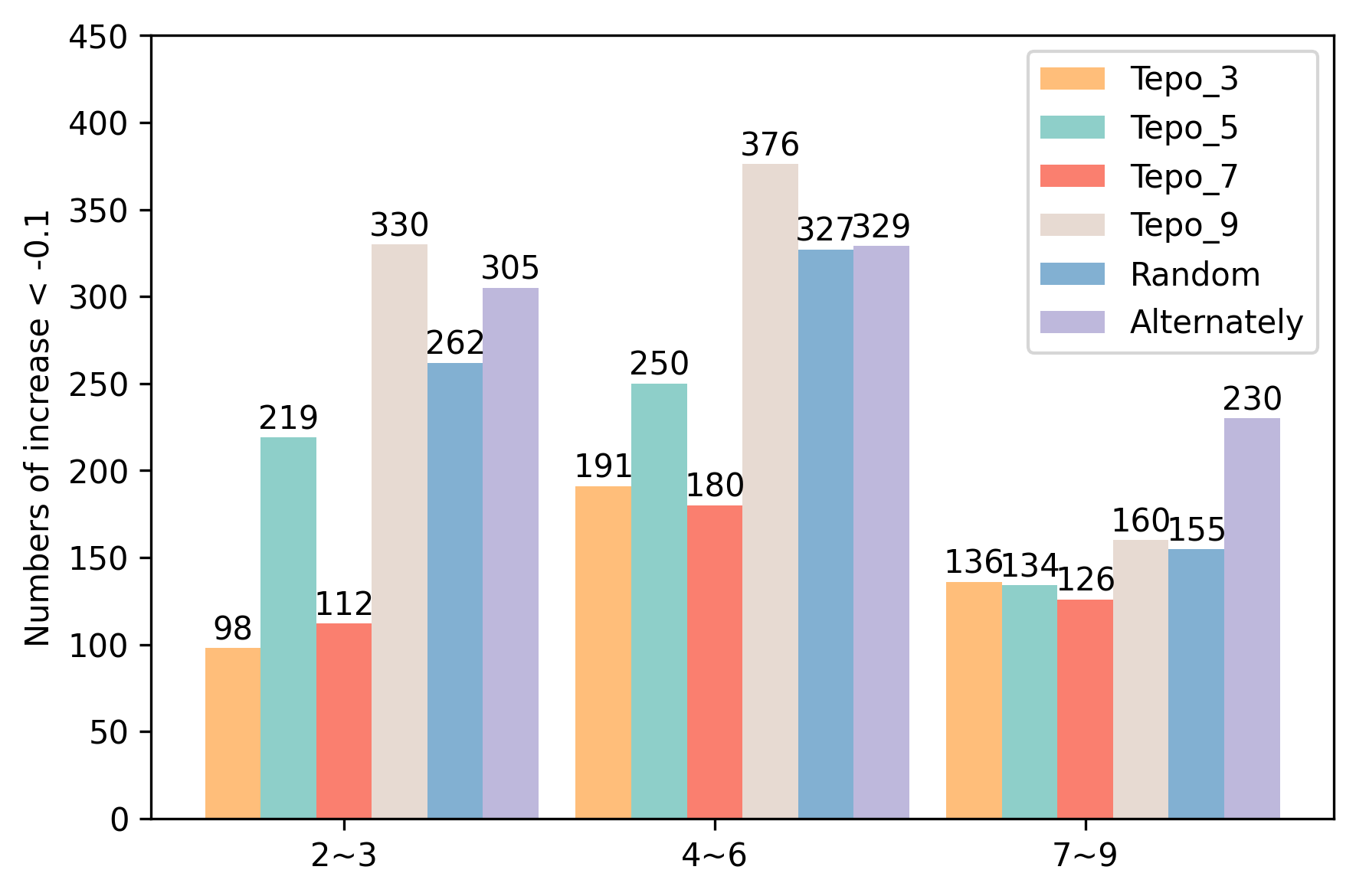}
    \caption{Comparison of the number of cases with increased dice score less than $-0.1$.}
    \label{fig:misunderstand}
\end{figure}

\paragraph{Q\#d: How stable are the strategies learned by TEPO?}
One issue that may affect the performance of TEPO is interactive misunderstanding, where user interactions result in reduced segmentation dice scores.
In this study, we consider interactive misunderstandings when the segmentation dice score decreases by over $0.1$.
We analyze the occurrence of interactive misunderstandings for different strategies on our test data, as presented in Tables \ref{tab:tepo_tab}.
For a more intuitive comparison, we plot the number of interactive misunderstandings for each strategy at different interaction steps in Figure \ref{fig:misunderstand}.
As TEPO-$2$ only applies to the initial two interactions, we exclude it from the plot.
Resultantly, the findings indicate that TEPO-$3$, TEPO-$5$, and TEPO-$7$ exhibit fewer misunderstandings than the random and alternating agents, thereby indicating superior stability and performance.

\subsubsection{Qualitative experimental analysis}
To evaluate the effectiveness of different strategies and investigate the causes of misunderstandings that occur with SAM, we conducted a qualitative analysis and present their performance on a single medical image in Figure~\ref{fig:case_study}.
The first column displays the raw image, while the middle columns show the interaction processes and corresponding segmentation outcomes. The last column provides the ground truth.
Among the different strategies, TEPO-$2$ demonstrates relatively weak performance, as it only involves two effective interactions.
The TEPO-$9$ and alternately changing agent purely used point-based interaction. TEPO-$9$ produces an equally good final outcome compared to the strategy with the bounding box and obtains the best result finally after nine interactions, the alternately changing policy performs poorly due to less effective interactions.
TEPO-$3$, and TEPO-$7$ consistently use the bounding box in the second interaction and select center points in all other interactions.
TEPO-$5$ uses the bounding box in the fourth interaction and center points in all other interactions. These three policies produce similar final results.
In addition, we observe a misunderstanding issue in some interactions, such as the seventh interaction with TEPO-$5$, and the eighth interaction with TEPO-$3$ and TEPO-$7$.
This is likely due to the corresponding region being too small for SAM to adequately understand the human feedback.

Overall, our results suggest that SAM cannot accurately achieve segmentation in a single interaction in medical tasks without being properly tuned. 
However, with multiple rounds of interaction, it can achieve considerable results. 
Moreover, the strategies learned by TEPO demonstrate better segmentation performance compared to rule-based strategies.

\section{Conclusion}
This paper focuses on assessing the potential of SAM’s zero-shot capabilities within the interactive medical image segmentation (IMIS) paradigm to amplify its benefits in the medical image segmentation (MIS) domain.
We introduce an innovative reinforcement learning-based framework, \textit{temporally-extended prompts optimization} (TEPO), to optimize prompts that can enhance segmentation accuracy in multi-step interaction situations.
Our empirical study, conducted on the \texttt{BraTS2020} benchmark, highlights the prompt sensitivity of SAM and demonstrates that TEPO can further enhance its zero-shot capability in the MIS domain. 
Specifically, TEPO successfully reduces the incidence of interactive misunderstandings, thus improving segmentation accuracy and stability in medical images. 
These findings make a valuable contribution to the development of advanced MIS techniques, showcasing the potential efficacy of prompts optimization which expands the zero-shot capability of foundation models like SAM.
\bibliography{main}

\end{document}